%% file: main_tech_report.tex
\documentclass[twoside,twocolumn]{article}

\usepackage{blindtext} 

\usepackage[sc]{mathpazo} 
\usepackage[T1]{fontenc} 
\usepackage{microtype} 

\usepackage[english]{babel} 

\usepackage[hmarginratio=1:1,top=32mm,columnsep=20pt]{geometry} 
\usepackage[format=plain, small,labelfont=bf,up,textfont=it,up]{caption} 
\usepackage{booktabs} 

\usepackage{lettrine} 

\usepackage{enumitem} 
\setlist[itemize]{noitemsep} 

\usepackage{abstract} 

\usepackage{titlesec} 
\renewcommand\thesection{\Roman{section}} 
\renewcommand\thesubsection{\roman{subsection}} 
\titleformat{\section}[block]{\large\scshape\centering}{\thesection.}{1em}{} 
\titleformat{\subsection}[block]{\large}{\thesubsection.}{1em}{} 

\usepackage{fancyhdr} 
\usepackage{titling} 

\usepackage{hyperref} 

\usepackage{amsmath}
\usepackage{amsfonts}
\usepackage{amssymb}
\usepackage{amsthm}

\usepackage{latexsym}
\usepackage{enumerate}
\usepackage{amsmath}
\usepackage{amssymb}
\usepackage{amsthm}
\usepackage{mathtools}
\usepackage{url}
\usepackage{xspace}
\usepackage{soul}
\usepackage{todonotes}
\setuldepth{x} 
\usepackage{algorithmicx}
\usepackage[noend]{algpseudocode}
\usepackage{todonotes}
\usepackage{filecontents}
\usepackage{tikz}
\usepackage{pgf}
\usepackage{pgfplots}
\usepackage{longtable}
\usepackage{array}
\usepackage{colortbl}
\usepackage{apacite} 
\usepackage{booktabs} 
\usepackage{siunitx} 
\usepackage{pgfplotstable} 
\usetikzlibrary{arrows,decorations,backgrounds,automata,shapes,patterns,decorations,backgrounds,plotmarks,calc,fit}
\usepackage{graphicx}
\usepackage{enumitem}
\setlist{nolistsep}
\input{macros}

\pretitle{\begin{center}\Large\bfseries} 
\posttitle{\end{center}} 
\title{DRAFT: Numerical Integration and Dynamic Discretization in Heuristic Search Planning over Hybrid Domains} 
\author{%
\textsc{Miquel Ramirez} \\[1ex] 
\normalsize University of Melbourne \\ 
\normalsize \href{mailto:miguel.ramirez@unimelb.edu.au}{miguel.ramirez@unimelb.edu.au} 
\and
\textsc{Enrico Scala} \\[1ex] 
\normalsize Australian National University \\ 
\normalsize \href{mailto:enrico.scala@anu.edu.au}{enrico.scala@anu.edu.au} 
\and
\textsc{Patrik Haslum} \\[1ex] 
\normalsize  Australian National University \\ 
\normalsize \href{mailto:patrik.haslum@anu.edu.au}{patrik.haslum@anu.edu.au} 
\and
\textsc{Sylvie Thiebaux} \\[1ex] 
\normalsize  Australian National University \\ 
\normalsize \href{mailto:sylvie.thiebaux@anu.edu.au}{sylvie.thiebaux@anu.edu.au} 
}
\date{\today} 
\renewcommand{\maketitlehookd}{%
\begin{abstract}
	In this paper we look into the problem of planning over hybrid domains, where change can be both discrete and instantaneous,
	or continuous over time. In addition, it is required that each state on the trajectory induced by the execution of plans complies
	with a given set of global constraints. We approach the computation of plans for such domains as the problem of searching over a
	deterministic state model. In this model, some of the successor states are obtained by solving numerically the so-called \emph{initial
		value problem} over a set of ordinary differential equations (ODE) given by the current plan prefix. These equations hold over
	time intervals whose duration is determined \emph{dynamically}, according to whether zero crossing events take place for a set of
	invariant conditions. The resulting planner, FS+, incorporates these features together with effective heuristic guidance.  FS+
	does not impose any of the syntactic restrictions on process effects often found on the existing literature on Hybrid Planning. A key concept of
	our approach is that a clear separation is struck between \emph{planning} and \emph{simulation} time steps. The former is the time allowed
	to observe the evolution of a given dynamical system before committing to a future course of action, whilst the later is part of the 
	model of the environment. FS+ is shown to be a robust planner over a diverse set of hybrid domains, taken from the existing literature
	on hybrid planning and systems.
\end{abstract}
}

\begin{document}

\maketitle

\section{Introduction}
\label{section:Introduction}

\input{introduction}

\section{Example: Zermelo's Navigation Problem}
\label{section:Running_Example}
\input{zermelo}
\section{\FPDDL}
\label{section:Language}
\input{language}

\section{Semantics of \FPDDL}
\label{section:Semantics}
\input{semantics}

\section{Branching and Computing Successor States}
\label{section:Branching_And_Computing_Successor_States}
\input{branching}

\section{Searching for Plans}
\label{section:Search_And_Heuristics}
\input{search}

\section{Experimental Evaluation}
\label{section:Experiments}
\input{experimental}
\section{Discussion}
\label{section:Discussion}
\input{discussion}

\section*{Acknowledgements}
This work has been partially supported by ARC project DP140104219, ``Robust AI Planning for Hybrid Systems''. 


\bibliographystyle{apacite}
\bibliography{crossref,master_biblio}


\end{document}

%% file: macros.tex
\newtheorem{theorem}{Theorem}

\theoremstyle{definition}
\newtheorem{definition}[theorem]{Definition}

\newcommand{\tup}[1]{\langle #1\rangle}            

\newcommand{\propername}[1]{\text{\textsf{\small #1}}\xspace}

\newcommand{\PDDL}{\textsc{pddl}}
\newcommand{\PDDLplus}{\textsc{pddl}$+$}
\newcommand{\PDDLPlus}{\textsc{pddl}$+$}
\newcommand{\FPDDL}{$\smallint$-\textsc{pddl}$+$}
\newcommand{\ADL}{\textsc{adl}}
\newcommand{\FS}{\propername{FS}}

\newcommand{\TmLpSat}{\propername{TM-LPSAT}}

\newcommand{\FSPlus}{\propername{FS+}}

\newcommand{\UPMurphi}{\propername{UPMurphi}}
\newcommand{\Dino}{\propername{DiNo}}

\newcommand{\ENHSP}{\propername{ENHSP}}

\newcommand{\ode}{\textsc{ode}}

\newcommand{\PNE}{\propername{PNE}}
\newcommand{\Fstrips}{\textsc{FSTRIPS}}
\newcommand{\FOL}{\textsc{FOL}}

\newcommand{\SMTPlan}{\textsc{SMTPlan+}}

\newcommand{\prop}[1]{{\cal P}_{A}}
\newcommand{\matr}[1]{\ensuremath{\mathbf{#1}}}

\newcommand{\OpTop}{\propername{OpTop}}

\newcommand{\Reals}{\ensuremath{\mathbb{R}}}

\newcommand{\thereals}{\ensuremath{\mathbb{R}}}

\newcommand{\varmode}{\ensuremath{\textbf{ctl}}}
\newcommand{\litleft}{\ensuremath{\text{left}}}
\newcommand{\litstraight}{\ensuremath{\text{straight}}}
\newcommand{\litright}{\ensuremath{\text{right}}}

\DeclarePairedDelimiter\floor{\lfloor}{\rfloor}

\newcommand{\VAL}{\propername{Val}}

%% file: introduction.tex
A central research topic in domain--independent automated planning is that of seeking plans
over \emph{hybrid} domain descriptions that feature both discrete and numeric state variables, as
well as discrete instantaneous and continuous durative change via actions and processes~\cite{fox:pddl_plus}.
A purely continuous dynamical system is defined by a set differential equations (\ode's)
that specifies how the system evolves over time~\cite{scheinerman:96:book}.
%
%
%
%
Hybrid planning problems correspond to control of \emph{switched dynamical systems},
which are driven by different dynamics (set of \ode's) in different \emph{modes}.
A mode can be defined by the values of discrete state variables, a region of the continuous
state space, or a combination of both~\cite{goebel:09:hybrid_dynamical_systems,ogata:control}.
Planning languages, such as \PDDLplus\ and our extension of \Fstrips, model switched dynamical
systems compactly in a factored way, avoiding the explicit enumeration of modes.


This paper introduces a heuristic search hybrid planner, \FSPlus.
%
%
%
%
Like McDermott's~\citeyear{mcdermott:03:icaps} hybrid planner, \OpTop, ours branches
over the set of applicable instantaneous actions plus a special ``waiting'' action $sim$,
that simulates continuous state evolution with the passing of time.
The duration of the waiting action, that discretises time, is not fixed to a single value or a suitably chosen set
like in~\cite{fox:12:batteries}, but rather \FSPlus~decides to use a smaller one than the
%
%
%
%
%
%
%
initially set \emph{planning} time step, $\Delta_{max}$.
The successor state that results from applying the waiting action is the result of simulating
system evolution, according to the dynamics of its current mode, for the duration
of the step; computing it is known as the \emph{initial value problem} in control
theory~\cite{ogata:control}.
For general dynamics there is no analytical solution to this problem,
but \emph{approximate} solutions can be obtained with a variety of numerical
integration methods~\cite{butcher:08:numerical_methods}. Such methods apply
a finer discretisation, using a \emph{simulation} time step $\Delta z$.
Finally, a validation step verifies that the invariant condition of the mode
~\cite{howey:03:sigplan} remains true throughout each simulation step.
If it does not -- which we refer to as a \emph{zero crossing event}, following
Shin \& Davies \citeyear{shin:05:tmlpsat} -- the interval is cut short.
Thus, \FSPlus, in an adaptive and to a high degree, unsupervised manner, breaks the time line
around specific time points, or \emph{happenings}~\cite{fox:pddl_plus},
effectively determining the right discretisation at each point in the plan
on-line.
This contrasts with previous work on hybrid planning, which either performs
plan validation (i.e., checking for zero crossings) off--line~\cite{howey:05:ictai,dellapenna:09:icaps},
or is restricted to specific classes of \ode's~\cite{shin:05:tmlpsat,lohr:12:mpc,coles:14:icaps,bryce:15:aaai,cashmore:16:icaps}.
A second feature is that \FSPlus~examines modes of the hybrid system only
as they are encountered in the search, in a manner similar to
Kuiper's~\citeyear{kuipers:86:qsim} qualitative simulation.
Since planning languages can compactly express hybrid systems with an
exponential number of modes as combinations of processes,
it is crucial to avoid generating or analysing all modes up front
\cite{lohr:12:mpc}.

The paper starts by illustrating a classical problem in control theory that motivated this 
research. After that, we introduce the language supported by \FSPlus, very similar to
\PDDLplus, but more succinct, that extends recently revisited classical
planning languages~\cite{frances:15:icaps}. Then the contributions of this paper are presented. 
First, we present the semantics of our planning language, that departs 
from~\cite{fox:pddl_plus} in some important aspects. Second, we show how 
a deterministic state model can account for the semantics given for hybrid planning,
describing the role played by numerical integration and the on--line detection of
zero crossing events. Third, we discuss very briefly how we integrate two recent 
 heuristics to construct  $h_{\FSPlus}$, a novel heuristic, \FSPlus~uses to guide the search
 for plans. The first of these heuristics is the Interval-Based Relaxation heuristic for classical numeric and hybrid planning~\cite{scala:16:ecai}, the other is the Constraint Relaxed Planning Graph heuristic developed 
 by~\cite{frances:15:icaps}.
%
%
%
%
%
Last, we discuss the performance of \FSPlus~over a diverse set of benchmarks featuring both with linear
and non--linear dynamics, and compare \FSPlus~with hybrid planners that can handle
such a diverse range of problems. We finish discussing the significance of our results and future work.

%
%
%
%
%

%% file: zermelo.tex
\begin{figure*}[ht!]
	\centering
	\includegraphics[width=0.45\linewidth,keepaspectratio]{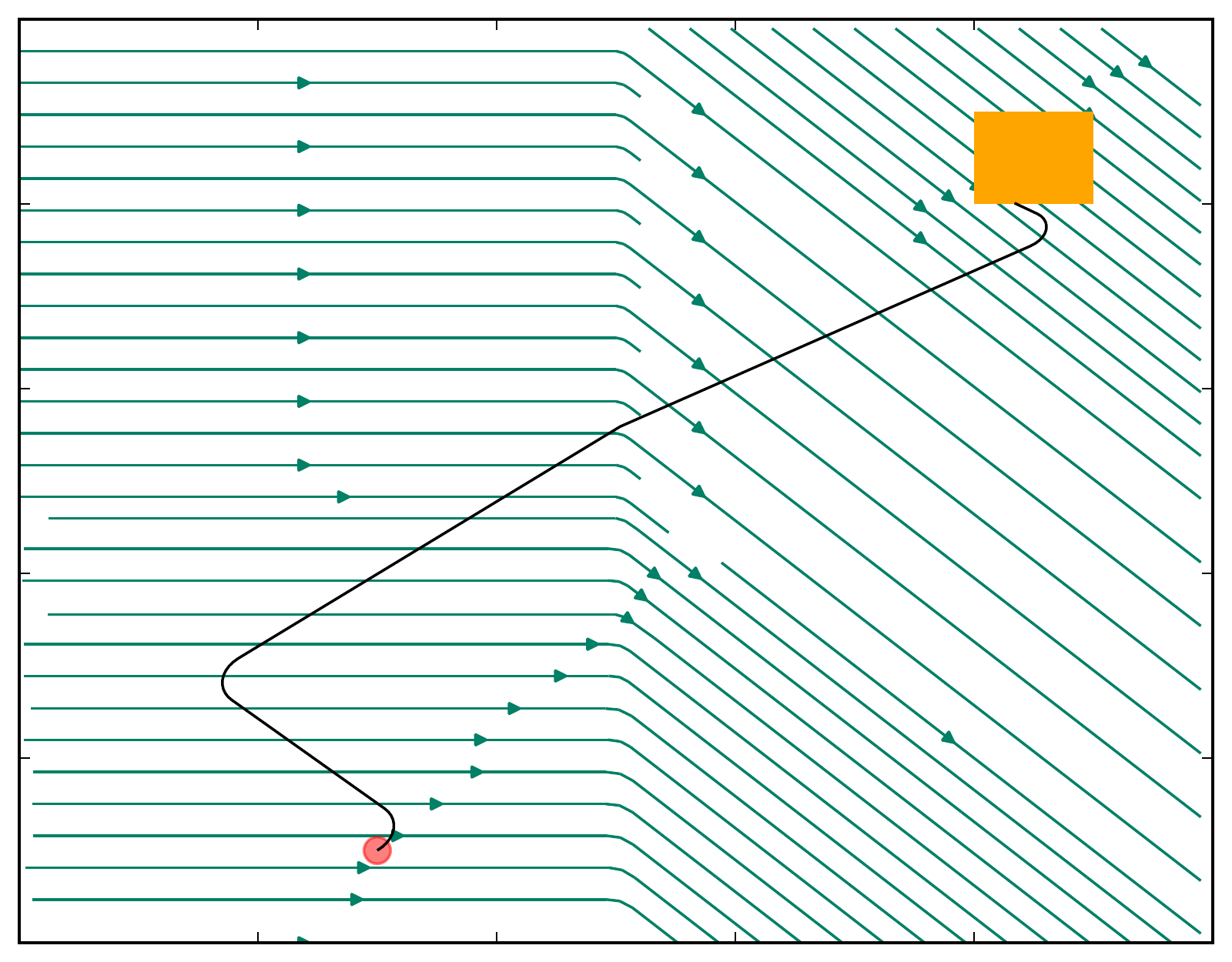}      
	\includegraphics[width=0.45\linewidth,keepaspectratio]{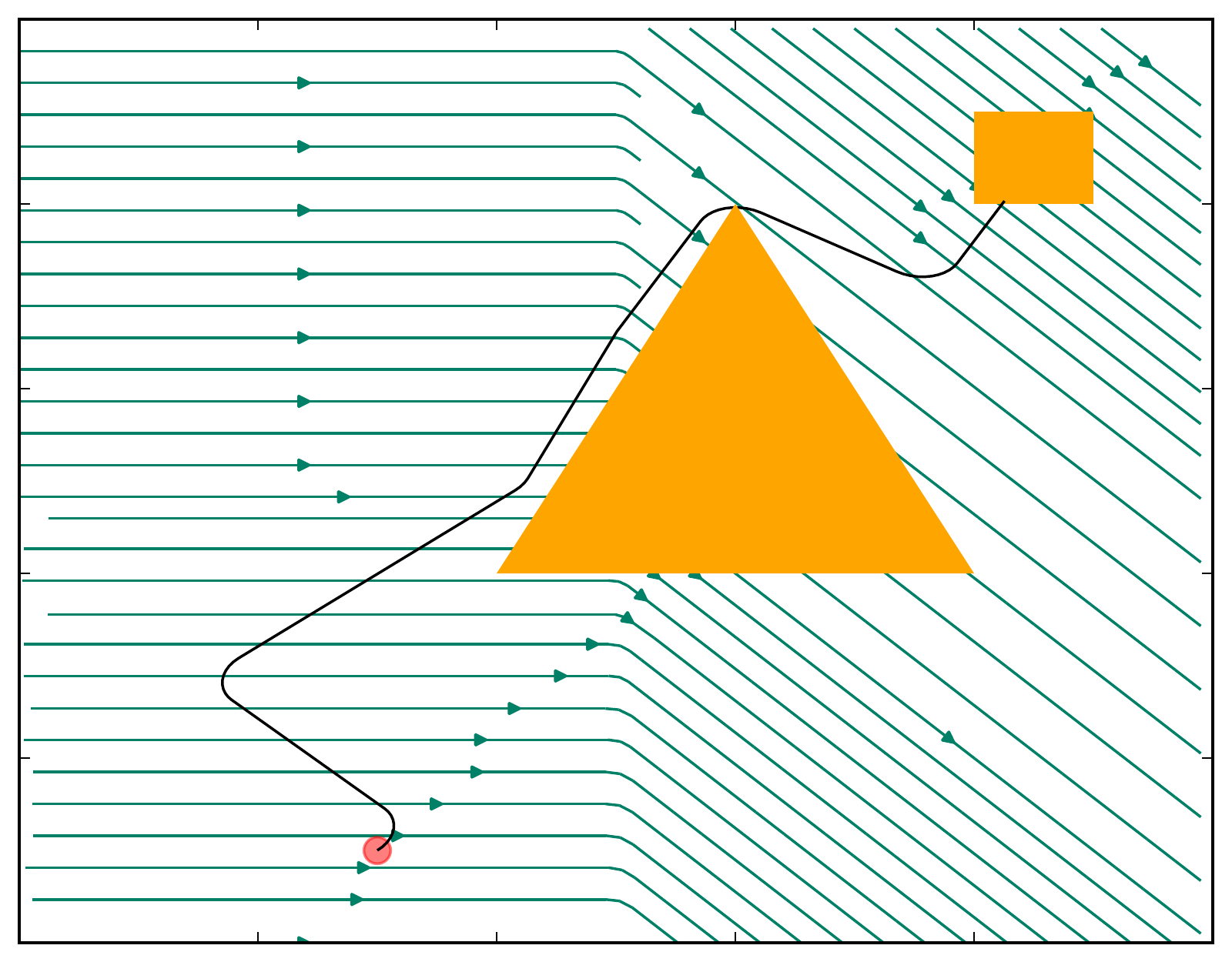}
	\caption{Trajectory found by \FSPlus~for two instances of the Zermelo navigation problem with non-homogeneous
		wind conditions (arrows show wind direction), and non--convex constraints in the instance on
		the right. The circle denotes the initial state $s_0$, the box   the set of goal states $s_G$ and the triangle is constraint.}
	\label{fig:zermelo}
\end{figure*}

Zermelo's navigation problem, proposed by Ernst Zermelo~\citeyear{zermelo:31:navigation}, is a classic 
optimal control problem that deals with a boat navigating on a body of water, starting from a point 
%
%
$s_0$ to end up within a designated goal region $s_G$. While simple, it has a vast number of 
real-world applications,
%
%
%
such as planning fuel efficient routes for commercial aircraft~\cite{soler:10:aircraft_trajectory}. 
The boat moves with \textit{speed} $v$,
its agility is given by the \textit{turning rate} $\rho$; both remain constant over time. It is desired to reach $s_G$ in the least possible time, yet the ship 
has to negotiate variable wind conditions, given by the position dependant drift vector $\matr{w}(x,y)$ $=$ $\langle u(x,y)$, $v(x,y) \rangle$. States consist of
three variables, the current location of the boat $(x,y) \in \Reals^2$ and its heading 
$\theta$\footnote{All state variables depend on the time $t$, that is, are subject to exogenous continuous change over variable $t$, that tracks the passage of time.}. For fixed heading $\theta$, we have that the location of the boat changes according to the \ode:
%
%
%
%
%
%
%
\begin{align}
\dot{x} & = v\, cos\, \theta+ u(x,y) \\
\dot{y} & = v\, sin\, \theta + v(x,y)
\label{eq:zermelo_dynamics}
\end{align}
The agent can steer the boat towards goal states $s_G$
\[
s_G \models x_{G}^{min} \leq x \leq x_{G}^{max} \land y_{G}^{min} \leq y \leq y_{G}^{max}
\]
by altering the angle $\theta$ in a
suitably defined manner. 
%
%
%
%
The function of $\theta$ over time is in effect the \emph{control signal}~\cite{ogata:control} for this
dynamical system. In this paper, we account for the range of possible signals by having \emph{instantaneous} actions to switch on, off or
modulate continuous change over certain state variables. In this example, the agent 
has available
three instantaneous actions $ahead$, $port$, $starboard$ that account respectively for keeping the boat rudder steady, push it towards the right, and to the left. Each of these actions sets an \emph{auxiliary}
variable \varmode~to the values \litstraight, \litleft~and \litright. These instantaneous actions cannot be
executed in any order, so we impose a further (logical) restriction by requiring to keep the rudder straight before being able to push it either towards the left and the right. These actions account for the
possible \emph{control switches} connecting the set of \emph{control modes} of a Hybrid 
Automata~\cite{henzinger:00:hybrid_automata}.
The variable $\varmode$ is then used to define the rate of change of $\theta$ in the following
manner: $\dot{\theta}$  $=$ $-\rho$ when $\varmode$ $=$ $\litleft$, 
$\dot{\theta}$  $=$ $\rho$ when $\varmode$ $=$ $\litright$, and $\dot{\theta}$ $=$ $0$
otherwise.
We note that angles are given in \emph{radians}.
\begin{table}[ht!]
	\begin{center}
		\footnotesize
		\begin{tabular}{l@{\hskip 3pt}l@{\hskip 3pt}l@{\hskip 3pt}l@{\hskip 3pt}l@{\hskip 3pt}l@{\hskip 3pt}l}
			\hline
			Time(secs)& 0& 100& 500& 600& 2030& 2130 \\
			\hline
			Action& $starboard$& $ahead$& $port$& $ahead$& $starboard$& $ahead$\\
			\hline
		\end{tabular}
		
	\end{center}
	\caption{{\footnotesize 
			Plan for the left image on Figure~\ref{fig:zermelo}, reaching goal area in $2,430$ seconds of simulated time, that required $0.32$ seconds
			of real time to be computed. $\Delta_{max}$ was set to $100$ seconds, and
			 a zero crossing event was
			detected at $t=2,030$.
			}}
	\label{table:zermelo_plan}
\end{table}
On the left of Figure~\ref{fig:zermelo} we can see the trajectory following from the plan in 
Table~\ref{table:zermelo_plan}. The rudder is left alone for quite some time,
until the boat almost sails past the goal, and then goes upwind towards $s_G$. Global constraints in that scenario require to keep the boat
within the bounding box, that coincides with the image extent. On the right in Figure~\ref{fig:zermelo} 
we can see the path that complies with an additional constraint: to stay outside of the golden triangle. 

%% file: language.tex
The planning language we will use in this work, \FPDDL\footnote{The $\smallint$  in \FPDDL~accounts both
for the role that the specific integration methods play as a implicit part of the modeling and also
for the use of functions. The latter follows from
the historical fact that in the 16th century, when calculus was being invented by Newton and Leibniz,
written English and German did not differentiate between the sounds for 's' and 'f'.},
%
%
is the result of integrating Functional STRIPS (\Fstrips)~\cite{geffner:00:fstrips}, 
and specific fragments of \PDDL~2.1 Level 2~\cite{fox:03:pddl} 
and \PDDL~2.1 Level 5~\cite{fox:pddl_plus}, also known as \PDDLplus.
\Fstrips~is a general modeling language for classical planning based on
the fragment of First Order Logic (\FOL)~involving \emph{constant},
\emph{functional} and \emph{relational} symbols (predicates), 
but no variable symbols, as originally proposed by Geffner, and
recently augmented with support
for quantification and conditional effects~\cite{frances:16:ijcai},
thus becoming practically equivalent to the finite--domain fragment
of \ADL~\cite{pednault:86:adl}. Its syntax is essentially the same as that of
the ``unofficial'' revision of \PDDL, \PDDL~3.1, first proposed by 
Helmert~\citeyear{helmert:08:pddl31} as the official \PDDL~variant
of the 2008 International Planning Competition, and fully formalised later by
Kov\'{a}cs~\citeyear{kovacs:11:pddl31}.
To this we have added support for features proposed in \PDDL~2.1 Level 2~\cite{fox:03:pddl} 
%
%
%
%
%
 to handle arithmetic expressions,  arbitrary algebraic and trigonometric functions~\cite{scala:16:ecai}, and  the notion of autonomous processes in \PDDL~2.1 Level 5~\cite{fox:pddl_plus} 
 to account continuous change over time.
  The implementation of \FPDDL~is built on top of
  that in the recent classical planner 
  \FS~\cite{frances:16:ijcai}.

States, preconditions and goals in \FPDDL~are described using fluent symbols, whose denotation changes as a result of
doing actions or the natural effect of processes over time. 
Those 
symbols whose denotation does not change are \emph{fixed} symbols and  include finite sets of object names, integer and real constants,
the arithmetic operators '$+$', '$-$', '$\times$', '$\div$', exponentiaton,
n--th roots, $sin()$, $cos()$ and $tan()$, 
as well as the relational symbols '$=$', '$>$', '$<$', '$\geq$' and '$\leq$', all
of them following a standard interpretation. Terms, atoms and formulas 
are defined from constant, function and relation symbols, with both terms
and symbols being \emph{typed}. Types are given by finite sets
of fixed constant symbols\footnote{Note that whenever we use the term 
``real numbers'' we actually refer to the finite set of rational numbers
that can be represented with the finite precision arithmetic supported in
most general--purpose programming languages. Similarly for ``integers''.}.
Terms $f(\alpha)$, where $f$ is a fluent symbol and $\alpha$ is a tuple of fixed
constant symbols, are called \emph{state variables}, and states are
determined by their values. Primitive Numeric Expressions 
(\PNE)~\cite{fox:03:pddl} correspond exactly with well formed arithmetic 
terms combining constants, state variables, arithmetic operators and other 
functions.
%
%
%
%

Instantaneous actions $a$ and processes $p$ are described by the type of
%
%
%
%
%
their arguments and two sets, the (pre)condition and the effects. Action $a$'s preconditions
$Pre(a)$ and process $p$'s conditions $K(p)$ are both formulae. 
%
%
%
%
%
%
%
Actions and processes 
%
%
%
%
%
differ in the definition of their effects. Action $a$ effects
update \emph{instantly} the denotation of state variables $f(\alpha)$ as a result of applying $a$. Process $p$ effects describe how the denotation of state variables $f(\alpha)$ changes as time goes by. State variables $f(\alpha)$ that appear \emph{only} in the left hand side 
of action effects are effectively \emph{inertial fluents}~\cite{gelfond:98:action}. On the other hand, those variables $f(\alpha)$ appearing on the
left hand side of process effects, and possibly as well in the effects of instantaneous
actions, are \emph{non--inertial}~\cite{giunchiglia:98:aaai} as their denotation can change even when left
alone.
Formally, the effect of an action $a$ is a set of updates of the form $g(\beta) := \xi_{a}$, where $g(\beta)$ and $\xi_a$ are terms of the same type, expressing how $g(\beta)$ changes when $a$ is taken. 
%
%
%
%
The effects of a process $p$ is also a set of update rules, but process updates, instead of
%
%
%
%
%
%
%
an assignment, are \ode's~$\dot{f}(\alpha) = \xi_{p}$\footnote{$\dot{f}(\alpha)$ denotes the
	derivative of $f(\alpha)$ over time.}.
The value of a time-dependent state variable $f(\alpha)$ after $t' - t_0$ time units, provided that $p$ is the \emph{only} process affecting $f(\alpha)$ as prescribed by the update rule
$\dot{f}(\alpha) = \xi_{p}$, is given by the following \emph{integral equation}
\begin{equation}
f'(\alpha) =  f_0(\alpha) + \int_{t_0}^{t'} \xi_{p}\, \partial t
\label{eq:integral_equation_single_var}
\end{equation}
where $t'$ $>$ $t_0$ is a positive finite number, $f'(\alpha)$\footnote{We decouple variable symbols from
	time $t$ following the notation typical from
%
%
%
%
%
 contemporary manuals on \ode's and numerical analysis e.g. Butcher's \citeyear{butcher:08:numerical_methods} .} is the value of state variable $f(\alpha)$ at time $t'$, and $t_0$ is the time associated with the \emph{initial conditions} $f_0(\alpha)$. The type allowed for $f(\alpha)$ and $\xi_{p}$ is restricted to be the real numbers. We impose a further
restriction on $\xi_{p}$, namely that it needs to be an \emph{integrable} function in the 
interval $[t_0,t']$, so the rightmost term in Equation~\ref{eq:integral_equation_single_var}
 is \emph{finite}. 
%
%
%
%
%

Global (state) constraints ${\cal C}$~\cite{lin:94:state_constraints} allow to describe compactly restrictions on the values that state variables $f(\alpha)$ can take over time. \FSPlus~ currently supports global constraints ${\cal C}$ given as 
%
%
%
%
%
%
%
%
%
CNF formulae, where each clause
$\varphi$ is a disjunction of relational formulae. 
This allows us to model state constraints similar to those proposed~\citeyear{ivankovic:14:icaps}.
%
%
%
%
%
Since \FPDDL~supports disjunctive formulae, by extension it accounts for implication as required by Ivankovic's \emph{switched} constraints $P \supset S$, where $P$ is
a conjunction of literals of so--called \emph{primary} variables, equivalent to our state variables $f(\alpha)$, and $S$ is an arbitrary
formula over so--called \emph{secondary} variables. However \FPDDL~cannot represent the latter, as their denotation in any given state is not fixed, as their value is given by those featured in the \emph{models} (satisfying assignments over $S$) of the constraints.

Last, the planning tasks we consider are tuples ${\cal P}$ $=$ $\tup{F,$ $I,$ $O,$ $P,$ $G,$ ${\cal C}}$ where $I$ and $G$
%
%
%
%
%
are the \emph{initial} state and \emph{goal} formula, 
$O$ is a set of instantaneous actions, $P$ is a set
of processes, ${\cal C}$ is a set of state constraints and $F$ describes the fluent symbols and their types. $I$ must define a \emph{unique} denotation for each of the symbols in $F$, and satisfy every $C \in {\cal C}$.

%% file: semantics.tex
We first briefly review the semantics of FSTRIPS, following Franc\`{e}s and
Geffner~\citeyear{frances:15:icaps,frances:16:ijcai}. Then we discuss
continuous change on state variables as 
the timeline is broken into a finite set of intervals, each with 
an associated set of \emph{active} processes~\cite{fox:pddl_plus}, or \emph{mode}~\cite{ogata:control}.

The logical interpretation of a state $s$ is described as follows in a bottom--up fashion. The denotation of a symbol or term $\phi$ in the state $s$ is written as $[\phi]^s$. The denotation 
 of objects or constants symbols $r$, is fixed and independent from $s$. 
%
%
%
%
Objects $o$ denote themselves and the denotation of constants (e.g. $3.14$) is given by the underlying programming language\footnote{In our case, C++.}. 
The denotation of fixed (typed) function and relational symbols can be provided 
\emph{extensionally}, by enumeration in $I$, or \emph{intensionally}, by attaching 
external procedures~\cite{dornhege:12:semantic}\footnote{Arithmetic, algebraic and relational symbols are represented intensionally when the types of state 
variables and constants present in an arithmetic term are the ``integers'' or the ``reals''.}.
The dynamic part of states $s$ is represented as the value of a finite set
of state variables $f(\alpha)$. From the fixed denotation of constant symbols
and the changing denotation of fluent symbols $f$ captured by the values $[f(\alpha)]^s$,
the denotation of arbitrary terms, atoms and formulas follows in a standard way.
An instantaneous action $a$ is deemed 
applicable in a state $s$ when $[Pre(a)]^{s}$ $=$ $\top$, and the state $s_a$ resulting from applying $a$ to $s$ is such that, 1) all  state constraints are satisfied, $[C]^{s_a}$ $=$ $\top$ for every state  constraint $C \in {\cal C}$, and 2) $[g(\beta)]^{s_a} = [\xi_a]^s$ for every update $g(\beta) := \xi_{a}$ triggered by $a$, and $[f(\alpha)]^{s_a}$ $=$ $[f(\alpha)]^s$ otherwise. 
 A sequence of instantaneous actions $act$ $=$ $(a_1,$ $\ldots,$ $a_j,$ $\ldots,$ $a_n)$, 
 $a_j \in O$, is \emph{applicable} in a state $s$ when $[Pre(a_1)]^s$ $=$ $\top$, and for every 
 intermediate state $s_{a_j}$, $j > 1$, resulting from applying action $a_{j-1}$ in $s_{a_{j-1}}$,
 it also holds that $[Pre(a_j)]^{s_{a_j}}$. We write  $s[act]$ for the state resulting from 
 applying sequence $act$ on state $s$.
%
%
%
%
%

Because our time domain, $\thereals^{+}_{0}$, is dense, the number of states is infinite.
%
%
%
%
%
We introduce structure into this line by borrowing 
Fox \& Long's~\citeyear{fox:pddl_plus} notion of 
\emph{happenings} $\eta$, distinguished points on the time line where an \emph{event} takes place or an instantaneous
action is executed. In between each pair of happenings is a \emph{steady interval} during which the state evolves
continuously according to the set of active processes, or modes, that characterize a dynamical system.
We require happenings to exist at any point where processes start or end, thus ensuring the set
of active processes is constant throughout each interval.

Formally, a happening $\eta$ is characterised by its \emph{timing}, $T(\eta)$, mapping $\eta$ to
$t \in \thereals^{+}_{0}$, the \emph{state} $s(\eta)$ at that time, and a sequence $act(\eta)$ of instantaneous
actions applied at the happening. Note that $s(\eta)$ is the state \emph{before} $act(\eta)$ is applied.
An \emph{interval} is characterised by the two happenings that mark its start and end: $H_i = \tup{\eta^{-}_{i}, \eta^{+}_{i}}$, where the end happening has no associated actions,
i.e., $act(\eta^{+}_{i}) = \langle \rangle$.
The duration of $H_i$ is the difference $\Delta_i = T(\eta^{+}_{i}) - T(\eta^{-}_{i})$
and we assume a finite upper and lower bounds on the duration of each interval, i.e., $\Delta_{min}$ $\leq$ $\Delta_i$ $\leq$ $\Delta_{max}$,
are provided as part of the problem description. The lower bound $\Delta_{min}$ can indeed be set to $0$,
but in that case we observe that introduces the possibility of plans with infinite length, their execution
being referred to as a \emph{Zeno's execution} in existing literature in hybrid control theory~\cite{goebel:09:hybrid_dynamical_systems}.
%
%
%
%
The values of state variables $f(\alpha)$ at the start of the interval $H_i$ are given by
%
%
%
%
%
the state $s^{i}_{0}$ that results from applying the sequence of actions $act(\eta^{-}_{i})$ associated
with the start happening to the state $s(\eta^{-}_{i})$, i.e. $s^{i}_{0} \equiv s(\eta^{-}_{i})[act(\eta^{-}_{i})]$.
During the interval $H_i$, continuous state variables may be affected by the set of processes
that are \emph{active} in $H_i$. The state at the end of $H_i$, $s(\eta^{+}_i)$, is defined by integrating
the active processes' effects, following the general form of Equation~\ref{eq:integral_equation_single_var}.
We define the set of active processes $\mu_i \subseteq P$, or \emph{mode}, associated with interval $H_i$, as
those whose conditions hold in the state at the interval's beginning:
\[
\mu_i = \{ p \,\mid\, p \in P, \,\, s^{i}_{0} \models  K(p) \}
\]
\noindent  We note that $\tup{s^{i}_{0},\mu_i}$ is the \emph{dynamical system} associated to $H_i$. Several active processes can affect the same state variable $f(\alpha)$.
Recall that process effects specify the rate of change:
we follow the standard convention that process effects superimpose, by adding
together the rates of change of all active processes affecting $f(\alpha)$
\cite{ogata:control}.
The value of state variable $f(\alpha)$ in state $s(\eta^{+}_i)$ is then given by
\begin{equation}
\label{eq:state_equation:processes_1}
[f(\alpha)]^{s(\eta^{+}_i)} = [f(\alpha)]^{s^{i}_{0}} +  \int_{T(\eta^{-}_i)}^{T(\eta^{+}_i)} \sum_{p \in {\cal R}_{f(\alpha)}} \delta_{p}^{i} \,\,\xi_{p} \,\,\partial t
\end{equation}
where ${\cal R}_{f(\alpha)} \subseteq P$ is the set of processes $p$ where $f(\alpha)$ appears on the left hand side of some effect $\xi_p$,
and the \emph{activation variable}, $\delta_p^i$, for each process $p$ is the \emph{characteristic function}
of the mode $\mu_{i}$, meaning that $\delta_{p}^{i} = 1$ if $p \in \mu_i$ and $0$ otherwise.
%
%
%
%
%
Equation~\ref{eq:state_equation:processes_1} can be simplified.
First, note that as long as $\mu_i$ remains \emph{stable},
that is, the set of active processes does not change over the duration of interval $H_i$,
$\delta_{p}^{i}$ does not depend on time $t$.
Second, the restriction 
on $\xi_{p}$ imposed in the previous Section, i.e. that $\xi_{p}$ is a continuous
function over $[T(\eta^{-}_i)$, $T(\eta^{+}_i)]$, enables the direct application of Fubini's Theorem~\citeyear{fubini:07:calculus}. Provided that $\mu_i$ is stable we can rewrite Equation~\ref{eq:state_equation:processes_1} as
\begin{equation}
\label{eq:state_equation:processes_2}
[f(\alpha)]^{s(\eta^{+}_i)} = [f(\alpha)]^{s^{i}_{0}} +   \sum_{p \in {\cal R}_{f(\alpha)} \cap \mu_i}  \int_{T(\eta^{-}_i)}^{T(\eta^{+}_i)} \xi_{p} \partial t
\end{equation}
\noindent Next, we define a condition that implies the interval $H_i$ is stable, i.e., that $\mu_i$
does not change during $H_i$.
To do this, we verify that 
none of the following happens at any point $t$ $\in$ $[T(\eta^{-}_{i}), T(\eta^{+}_{i})]$:
%
%
(1) the truth of the conditions $K(p)$ of some process $p \in P$ changes,
(2) some state constraint $C \in {\cal C}$ is violated, and
(3) we do not ``shoot through'' sets of states where the goal $G$ is true.
The absence of each of these events can be expressed as a condition,
the conjunction of these three conditions is the \emph{invariant} formula
\begin{equation}
\label{eq:steady interval_invariant}
{\cal I}(H_i) : 
\neg G \land \bigwedge_{p \in \mu_i } K(p) \land  \bigwedge_{p' \in P \setminus \mu_i} \neg K(p') \land \bigwedge_{C \in {\cal C}} C
\end{equation}
\noindent \cite{howey:03:sigplan,howey:05:ictai}.
If the invariant ${\cal I}(H_i)$ holds throughout $H_i$, then $H_i$ is stable.
Note that sub-formulae of ${\cal I}(H_i)$  that mention \emph{only} state variables
$f(\alpha)$ not affected by any process 
always remain true over $H_i$.
%
%
%
Following Shin \& Davies' \citeyear{shin:05:tmlpsat}, 
we refer to a change in the truth value of ${\cal I}(H_i)$ as a \emph{zero crossing event}.
\begin{definition}{(Zero Crossing Event)}
Let $H_i$ be a interval with timings $T(\eta^{-}_{i})$ and $T(\eta^{+}_{i}) $, and invariant
${\cal I}(H_i)$. A \emph{zero crossing event} occurs whenever for some
$t'$ s.t. $ T(\eta^{-}_{i})$  $<$ $t'$ $<$ $T(\eta^{+}_{i})$, ${\cal I}(H_i)$ is \emph{false}.
\end{definition}

\begin{definition}{(Steady Intervals)}
Let $H_i$ be a interval with timings $T(\eta^{-}_{i})$ and $T(\eta^{+}_{i}) $, state $s^{i}_0$ and 
mode $\mu_i$. Whenever \emph{zero} crossing event occurs for some $t'$ s.t. $ T(\eta^{-}_{i})$  $<$ $t'$ $<$ $T(\eta^{+}_{i})$, then the interval $H_i$ is a \emph{steady interval}, whose dynamics
are fully described by the dynamical system $\tup{s^{i}_{0},\mu_{i}}$.
\end{definition}
We are now ready to define what is a solution for our planning tasks. A
 valid plan $\pi$ for hybrid planning task ${\cal P}$ is a finite sequence 
\[
\pi = (H_0, H_1, \ldots, H_i, \ldots, H_m)
\]
of steady intervals $H_i$, such that
(1) $s(\eta^{+}_{m}) \models G$, and 
(2) $T(\eta^{-}_{m}) = T(\eta^{+}_{i-1})$.
%
%
%
%
The \emph{optimality} of plans depends on
%
%
%
%
%
%
%
the \emph{metric} the modeller specifies for ${\cal P}$ as evaluated on $s(\eta^{+}_m)$: our extended \Fstrips~language,
like \PDDL~3.1, enables the modeller to specify both obvious metrics such as the overall duration
of $\pi$ (i.e. $\sum \Delta_i$)
and more intricate ones as needed. 

\subsection{Comparison with \PDDLPlus}

The planning language we define is very similar to, and certainly owes many of
its core concepts to, the specification of \PDDLPlus~ given by Fox \& Long ~\citeyear{fox:pddl_plus}.
However, it separates in two ways, in the interpretation of plans and not accounting for the \PDDLplus~notion
of \emph{events} completely.

First, we consider only the computation of plans with finite duration and representation,
consisting of a finite but unbounded number of happenings. Our interpretation
of happenings is that they break the continuous timeline into a finite sequence of
intervals during which the continuous effects on state variables are \emph{stationary}.
The number of states in each interval is infinite, but all such intermediate states are
%
%
%
%
%
%
%
cataloged \emph{implicitly} by the states at the happenings that mark the extent of the
interval~\cite{pednault:86:adl}.

Second, we allow the planner to execute a finite but unbounded sequence of instantaneous
actions at a happening $\eta$ without any restriction, such as commutativity, on their effects.
This allows us to model, for instance, the network of valves in the rocket engine discussed in the classic work of Williams \& Nayak~\citeyear{williams:96:aaai} without an explosion in the number of instantaneous actions required to represent all possible combinations of valve positions.
In contrast, \PDDLPlus~ mandates a non-zero time separation (known as the ``$\epsilon$'')
between non-commutative instantaneous actions. This restriction is motivated by the
assumption that actions, although modelled as instantaneous, cannot actually be executed
in zero time \cite{fox:pddl_plus}.
However, for those cases in which a temporal separation, or ``cool down'' period, between
actions is motivated by the application domain, this can be modelled (using an auxiliary process
representing a timer, for instance) also in our setting.
Thus, no expressivity is lost when it comes to model limitations in the execution of plans.

Finally \PDDLplus~ considers \emph{events} to be ``first--class citizens'' in the language,
with semantics best described as ``exogenous'' instantaneous actions, rather than implicitly
defined as points where invariant conditions change.
Although events are in some cases a natural and convenient modeling device, some of their effects can
also be captured by introducing suitably defined global constraints or compiled into process 
preconditions. For instance in the \textsc{Mars Solar Rover} domain proposed by~\citeyear{howey:05:aaai},
one can do away with the events \texttt{sunset} and \texttt{sunrise} by having a global timer for the
whole day, and modifying accordingly the preconditions of the processes \texttt{day time} and 
\texttt{night time}.
This also avoids some of the problems Fox et al observe to be associated with plan validation in the presence of
events. On the other hand, uses of events to model spontaneous changes in the dynamics, are not accounted for. A natural and familiar example of  such phenomena is that of \emph{ellastic collisions} between bodies with the same mass, where the direction of acceleration changes as consequence of the collision. Accounting for these would be necessary to model accurately as a hybrid domain real-world tasks such as playing solitaire pool.

%% file: branching.tex
As outlined in the Introduction, \FSPlus~searches for plans $\pi$ of hybrid planning
task $P$ via forward search over a deterministic state model $M({\cal P}) = \tup{S$, $s_I$,
$A'$, $S_G$, $App$, $f$ $}$ where each element is defined as follows. The \emph{state space}
$S$ is given by all the possible combinations of denotations for fluents $F$ plus
an auxiliary state variable $t()$ to represent the location of states $s$ on the time line. Actions
$A' =$  $O$ $\cup$ $\{ sim  \}$ include the instantaneous actions $O$ in ${\cal P}$ and
a $sim$ action that updates $t()$ and simulates the world dynamics as given by the
%
%
%
%
%
continuous effects of processes $P$. Goal states $S_{G} \subseteq S$ are those states
$s$ s.t. $[G]^s=\top$; $s_I$ is like $I$.
The
applicability function $App$ corresponds exactly with the notion of applicable actions
discussed in the previous Section, while $sim$ actions can be applied in every state $s \in S$.  $f(a,s) = s_a$ for instantaneous actions $a$; we devote the 
rest of this Section to define $f(sim,s)$. 
Solutions to $M({\cal P})$ are paths $\pi'$ $=$ $(a_0$, $\ldots$, $a_n)$ connecting $s_I$ with some $s \in S_G$. The
%
%
%
%
plan $\pi$ made up of intervals $H_j$ is obtained from paths $\pi'$ by observing that (1) for every  action
$a_j$ $=$ $sim$ action in  $\pi'$,
%
%
%
%
%
 there is an interval $H_j$ in $\pi$, (2) the timing of happening $\eta^{-}_{j}$ is $[t()]^{s_l}$, where $s_l$ $=$ $s_0$ or $s_l$ = $f(a_l,s_{l-1})$,  $a_l$ $\in$ $\pi'$
 $a_l$ $=$ $sim$, $l  < j$, (3) the timing of happening $\eta^{+}_{j}$ is $[t()]^{s_j}$, $s_j$ $=$ $f(s_{j-1},sim)$, and (4) 
 $act(\eta^{-}_j)$ $\subset$ $\pi'$, defined as $act(\eta^{-}_j)$ $=$ $($ $a_{k}$, $\ldots$, $a_{j-1}$ $)$, $k=0$ or $a_{k-1}$ $=$ $sim$.
As suggested by this mapping of paths between initial and goal states in $M({\cal P})$
into sequences of intervals $H_i$, the action $sim$ (1) \emph{predicts successor
states} by solving 
Equation~\ref{eq:state_equation:processes_2} for every state variable $f(\alpha)$ appearing
%
%
%
on the left--hand side of effects of processes $p$ $\in$ $\mu_i$, and (2) 
\emph{validates} the assumption that $\mu_i$ is \emph{stable} checking whether for some $t_{zc}$  in the interval $[T(\eta^{-}_{i})$, $T(\eta^{-}_{i})$ $+$ $\Delta_{max}]$ the truth of ${\cal I}(H_{i})$ changes.
When that is the case, auxiliary variable $t()$ is set to $t_{zc}$ instead of $T(\eta^{-}_{i})$ $+$ $\Delta_{max}$, in turn setting $T(\eta^{+}_{i})$ to $t_{zc}$ as well. 

We note that  both of these problems cannot be solved \emph{exactly} in general, as both state prediction, or general symbolic integration, and validation, or finding roots of real--valued functions~\cite{howey:03:sigplan}, can be shown to be undecidable by Richardson's Theorem~\citeyear{richardson:68:undecidable}. 
%
%
%
%
Consequently, it will never be
possible to \emph{guarantee} that plans are valid, but that is not necessary
to compute plans that are accurate enough 
to inform the solution of real-world engineering problems. While
exact, general solutions are out of reach, we can turn to numerical
approximation methods for both prediction and validation. We discuss how
we approximate the prediction of successor states and their validation next.

\subsection{Related Work: Analytical Solutions and Linear Dynamics}
%
%
%
%
%

The implications of Richardson's Theorem on the validity of plans are indeed quite negative,
and it has motivated the planning community to look into less expressive, yet still powerful and
widely applicable, fragments of hybrid planning, where the form of processes effects is restricted
in some way. We devote this Section to briefly discuss existing work on a fragment that, while still
undecidable, does not require to explicitly discretise time.

A substantial part of the existing literature on hybrid planning studies domains where the right--hand side $\xi_{p}$ of processes $p$ in modes $\mu_i$ reachable from the initial state, and hence
the dynamical systems associated with such modes $\tup{s^{i}_{0},\mu_{i}}$, has a specific form:
that of general \emph{linear} expressions, 
\begin{align}
\xi_p & :\,\,  b_p + \sum_{f(\alpha) \in F} w_{f(\alpha)}^{p} f(\alpha) 
\label{eq:linear_process_pattern}
\end{align}
\noindent In that case, the combined effects of processes
in $\mu_i$ can be written compactly in matrix form as follows:
\begin{align}
\dot{\matr{x}} = \matr{A} \matr{x} + \matr{b}
\label{eq:linear_system_equation}
\end{align}
\noindent where $\matr{x}$ is made of (time--varying) state variables $f(\alpha)$ and both $\matr{A}$ and $\matr{b}$ 
follow from the coefficients in Equation~\ref{eq:linear_process_pattern}.
The dynamical systems Equation~\ref{eq:linear_system_equation} accounts for are a useful and
well known class of dynamical systems, known as Linear Time--Invariant (LTI) systems~\cite{scheinerman:96:book,ogata:control},
for which there exists an \emph{analytical solution} to Equation~\ref{eq:linear_system_equation}.  LTI systems are good models 
 for a huge range of domains~\cite{lohr:12:mpc}, yet still lack generality 
since linear approximations of physical processes are not always reasonable.
From a practical standpoint, solving the initial value problem amounts to
computing the \emph{exponential} of matrix $\matr{A}$. Matrix exponentiation is a non--trivial linear algebra problem~\cite{horn:matrix_analysis}, that needs to be solved for 
\emph{every} possible system $\tup{s^{i}_{0},\mu_i}$. Existing approaches rely on precomputing the
%
%
%
%
%
closed form for Equation~\ref{eq:linear_system_equation}, something which is tricky in domains like McDermott's \textsc{Convoys}~\citeyear{mcdermott:03:icaps}
where enumeration can be impractical.


\subsection{To Successor States via Numerical Integration}

Computing the state corresponding to the happening $\eta^{+}_i$ requires $sim$ to solve the \emph{initial value} problem~\cite{butcher:08:numerical_methods}, for the 
dynamical system $\tup{s^{i}_{0},\mu_i}$ associated with the interval $H_i$. \FSPlus~does so relying on numerical integration methods~\cite{butcher:08:numerical_methods} that do not require 
syntactic restrictions on the effects $\xi_p$ of processes $p \in P$ in order to be applicable. This generality comes at a cost: since numerical integration
relies on discretization of the free variable, time $t$ in this case, and we need to
introduce a new parameter, $\Delta z$, the \emph{simulation step}. The computation of the state
$s(\eta_{i}^{+})$ for a given interval $H_i$, proceeds as follows. We start observing that given 
any happening $\eta$, a mode $\mu_i$ and the state $s(\eta)$, the state $s(\eta')$ of another happening $\eta'$ s.t. $T(\eta')$ $=$ $T(\eta) + \Delta z$ is defined as
\[
s(\eta') = \Phi[s(\eta),\mu_i,\Delta z]
\]
where $\Phi[\cdot]$ is the specific numerical integration method being used to predict the values
of state variables $f(\alpha)$ being changed by processes $p \in \mu_i$. Computing the state
$s(\eta_{i}^{+})$ amounts to \emph{integrate} Equation~\ref{eq:state_equation:processes_2} over
intervals of duration $\Delta z$, and repeat this $k$ times where
\[
k = \floor*{\frac{T(\eta_{i}^{+}) - T(\eta_{i}^{-})}{\Delta z}}
\]
\noindent Whenever $k$ $\Delta z$ $<$ $\Delta_i$ an additional
%
%
%
%
%
call to the integration method $\Phi[\cdot]$ using $\Delta z'$ as the simulation step is needed and
%
%
%
%
%
\[
\Delta z' = T(\eta_{i}^{+}) - ( T(\eta_{i}^{-}) + k \Delta z)
\]
\noindent The simplest integrator $\Phi[\cdot]$ implemented in \FSPlus~is the Explicit Euler Method~\cite{butcher:08:numerical_methods}, that determines the values of
$f(\alpha)$ in state $s_k$ following Equation
\begin{equation}
[f(\alpha)]^{s(\eta')} = [f(\alpha)]^{s(\eta)} + \Delta z \sum_{p \in \mu_i} [\xi_{p}]^{s(\eta)}
\label{eq:explicit_euler}
\end{equation}
\noindent which is the \emph{recurrence relation} for integral Equation~\ref{eq:state_equation:processes_2}.
The convergence of numerical integration methods like the one in Equation~\ref{eq:explicit_euler} is very sensitive
to the choice of $\Delta z$, and for non--linear $\xi_p$ these methods can be easily shown to
diverge even for small values of $\Delta z$. On the other hand, more robust numerical integration methods, such 
as the Runge--Kutta integrators~\cite{butcher:08:numerical_methods}, are significantly more 
complicated than Equation~\ref{eq:explicit_euler} yet still much cheaper than matrix exponentiation.
Amongst these,
%
%
%
\FSPlus~currently implements the \emph{midpoint rule}, the 2nd order Runge--Kutta
integrator that Butcher refers to as $RK22$.
\FSPlus~also implements the \emph{iterative} or \emph{multi--step} Implicit Euler method
\begin{equation}
[f(\alpha)]^{s_{j+1}(\eta')} = [f(\alpha)]^{s(\eta)} + \Delta z \sum_{p \in \mu_i} [\xi_{p}]^{s_j(\eta')}
\label{eq:implicit_euler}
\end{equation}
\noindent where $[f(\alpha)]^{s_{0}(\eta')}$ is given by Equation~\ref{eq:explicit_euler}, and
iteration continues until the following fixed--point is reached:
\begin{equation}
|\,[f(\alpha)]^{s(\eta')}_{j+1} - [\xi_{p}]^{s(\eta')}_{j}\,| < \epsilon
\label{eq:implict_euler_fixed_point}
\end{equation}

Last, for problems where all process effects are linear expressions, we note that  
the ``messy'' numerical integrator  $\Phi[\cdot]$ currently available
in \FSPlus~could be readily substituted with the analytical solution of the LTI system 
given by $\mu_i$, calling an external solver \emph{online} instead of doing so as pre--processing
step, during the search. Such a change would 
entail significant gains in precision, but we suspect the cost of computing the analytical solution  would have 
a significant negative effect on run--times.

\subsection{Testing for Zero Crossings}


The invariant ${\cal I}(H_i)$ is a conjunction of parts, each of which can
be evaluated separately. Some parts may be disjunctive (thus non-convex)
as a result of negating conjunctive goals $G$ or process conditions $K(p)$
\cite{howey:03:sigplan}, or from disjunctive global constraints.
Recall, however, that the main aim of validation is to prove the absence of a
zero-crossing event in the simulated time interval. This allows us to simplify
the problem by testing \emph{necessary conditions} for a zero-crossing event to occur;
if those are not satisfied, we can conclude no such event happened, and hence
that the interval is steady. If a zero-crossing event may have occured, it is
sufficient to find an time point $t$ in the interval such that the event is
necessarily after $t$. The length of the steady interval is then shortened,
and the planner inserts a new happening at $t$ from which it can branch.

To validate a disjunctive formula $\varphi = \phi_1 \lor \ldots \lor \phi_l$
over an interval starting from state $s$,
we test instead $\varphi^{s} = \bigwedge_{j \text{s.t.} s \models \phi_j} \phi_j$, that is,
the conjunction of the disjuncts that are true in $s$, since the falsification of at
least one of those disjuncts is a necessary condition for $\varphi$ to become false.
%
%
%
%
%
%
After strengthening ${\cal I}(H_i)$ in this fashion, we introduce $k$ happenings $\eta_{i}^{j}$ with 
timings $T(\eta_{i}^{j})$ $=$ $T(\eta_{i}^{-})$ $+$ $j$ $\Delta z$, and $T(\eta_{i}^{-})$ $+$ $k$ $\Delta z$ $<$ $T(\eta_{i}^{+})$, and states resulting from calls to an integrator $\Phi[\cdot]$
using a \emph{finer grained} simulation step $\Delta h$\footnote{$\Delta h$ is set 
	somewhat arbitrarily to $0.1\Delta z$.}.
Where $\xi_{l}$, $\xi_{r}$ are arithmetic terms and $\oplus$ is a relational symbol, the truth of an \emph{atomic} formula $\psi = \xi_{l} \oplus \xi_{r}$, 
in state $s$ is tied to the \emph{sign} of the function $f_{\psi}$
\begin{equation}
f_{\psi}(s) =  [\xi_{l} - \xi_{r}]^{s}
\end{equation}
\noindent -- $\psi$ can only change from true to false if $f_{\psi}(s)$ changes sign.
Whether this occurs in the interval between two consecutive happenings $\eta_{i}^{j}$, $\eta_{i}^{j+1}$
can be determined
by direct application of the following classical result from mathematical analysis:
\begin{theorem}{(Intermediate Value Theorem)}
Let $f$ be \emph{continuous} on the closed interval $[a,b]$. If $f(a)$ $\leq$ $y$ $\leq$ $f(b)$ or
$f(b)$ $\leq$ $y$ $\leq$ $f(a)$ then there exists point $c$ such that $a$ $\leq$ $c$ $\leq$ $b$
and $f(c)$ $=$ $y$.
\end{theorem}
\noindent We observe that (1) $f_{\psi}$ is a continuous function, following from the constraint
on effects of processes $\xi_p$ to be \emph{integrable}, and (2) if
%
%
%
%
$f_{\psi}(\eta_{i}^{j})$ $<$ $0$ ($>$ $0$) and $f_{\psi}(\eta_{i}^{j+1})$ $>0$ ( $<$ $0$), then there necessarily exists a happening $\eta_{zc}$, $T(\eta_{zc}) \in [T(\eta_{i}^{j}),T(\eta_{i}^{j+1})]$ such that $f_{\psi}(\eta_{zc})$ $=$ $0$, i.e. a zero--crossing event for $\psi$ takes place at time $T(\eta_{zc})$.
The earlier strengthening of disjunctions implies that $\varphi^s$ is falsified as soon
as this happens for any atomic formula $\psi$ is.
In this case, the interval $H_i$ is shortened by setting $\eta^{+}_{i} = \eta^{j}_{i}$.

The treatment of disjunctive invariants means that happenings may be inserted
where not truly needed, since a disjunction can remain true even if one of its disjuncts
is falsified. This may lead to a deeper search than necessary, but any path found will
still be valid.
If the dynamics in the interval oscillate rapidly, the indicator function $f_{\psi}$ may
change sign several times within the validation time step $\Delta h$; thus, the procedure
may also fail to detect a zero-crossing event.
As noted earlier, this is an unavoidable consequence of the undecidability of the
root-finding problem in general.

%% file: search.tex
In the previous Section we have mapped our take on planning over hybrid systems into the problem of finding a path
in a deterministic state model $M({\cal P})$. This enables us to use, off--the--shelf without modifications,
any known blind or heuristic search algorithm, such as Breadth--First Search or 
Greedy Best First Search, to seek plans.
Still, heuristic guidance is essential for scaling up over these domains: none of the instances
discussed in the next Section are solvable by Breadth--First Search \emph{when} doing on--line
validation of successor states. In order to obtain a heuristic, \FSPlus~follows Scala et al.~\citeyear{scala:16:ecai}, and compiles away processes $p$
as instantaneous actions $a_p$. That is, for each update rule $\dot{f}(\alpha) := \xi_p$
%
%
%
%
%
we introduce an action effect that mimicks Equation~\ref{eq:explicit_euler}
\begin{align}
f(\alpha) :=  f(\alpha) + \Delta_{max}\, \xi_p
\end{align}
\noindent where $\Delta_{max}$ is the planning step. Then, on the resulting classical
%
%
%
%
planning task $P_{num}$, which is like ${\cal P}$ but with $P$ $=$ $\emptyset$, we apply a novel heuristic, $h_{\FSPlus}$, that integrates two recent works.
One is the \textsc{Aibr} relaxation~\cite{scala:16:ecai} that generalizes the notion of
%
%
%
%
%
\emph{value--accumulation semantics}~ \cite{gregory:12:icaps} to \emph{dense intervals}
defined over numeric variables. The other is the \textsc{Crpg} heuristic by Franc\`{e}s 
%
%
%
%
%
and Geffner~\citeyear{frances:15:icaps}, developed to exploit the structure of planning 
tasks exposed by \Fstrips. $h_{\FSPlus}$ extends the constraint language supported
by \textsc{Crpg} to support real--valued variables, defined over intensionally represented
 domains, i.e. intervals instead of sets. Scala et al 's \emph{convex union} operator is then used to succinctly accumulate the intervals revised
 after applying actions on each layer of the \textsc{Crpg}, thus \emph{propagating upper and lower bounds} over
 numeric variables \emph{along with} atomic formulae across layers. $h_{\FSPlus}$ estimate corresponds to the number of actions and processes that are needed to get into a layer of the \textsc{Crpg} where the goal is satisfied. \FSPlus~relies on the \textsc{Cp} framework \textsc{GeCode}\footnote{Version 5.0.0 available at \url{http://www.gecode.org/}} as the 
 latest versions of Franc\`{e}s and Geffner~\citeyear{frances:16:ijcai} planners do.

%% file: experimental.tex
\begin{table*}[t!]
	\begin{center}
		\footnotesize
		\begin{tabular}{l@{}|c@{ }|c@{ }|@{}c@{}}
			Domain & P \& G & Dyn & Source \\
			\hline
			\textsc{Agile  Robot} & L & L, H & \cite{lohr:14:dissertation} \\
			\textsc{Orbital  Rendezvous} & L & L, G & \cite{lohr:14:dissertation} \\
			\textsc{Convoys} & L & L,NH & \cite{mcdermott:03:icaps} \\
			\textsc{Intercept} & NL & L, NH & \cite{scala:16:ecai}\\
			\textsc{Zermelo} & L & NL & \cite{zermelo:31:navigation}\\
			\textsc{1D  Powered  Descent} & L & L, G & \citeyear{piotrowski:16:ijcai} \\
			\textsc{Non  Linear  Car} & L & NL & \cite{bryce:15:aaai} \\
			\textsc{Dino's Car} & L & NL & \cite{piotrowski:16:ijcai}
		\end{tabular}
	\end{center}
	\caption{{\footnotesize
			Taxonomy of domains according to expressions, in preconditions, process conditions, goals (P \& G)
			and process effects (Dyn), being linear (L) or non--linear (NL) . We further distinguish three different
			sub-clases for Linear dynamics: general (G), homogenous (H), non--homogeneous (NH). See text for details.
		}}
		\label{table:domain_taxonomy}

	\end{table*}
In order to evaluate \FSPlus\ we have selected a number of benchmarks already proposed in the literature on
hybrid and numeric planning, as well as the Zermelo's navigation problem (\textsc{Zermelo}) discussed in
Section II. The criterion used to select benchmarks was first and foremost to
consider a diverse representation of the kind of linear and non--linear that can be modeled with \FPDDL, and \PDDLPlus, with non--trivial branching factor. Table~\ref{table:domain_taxonomy} lists the domains considered, along with
the pointers to the publication describing them and a classification of their dynamics.
Benchmarks are considered to have \emph{linear} dynamics (L) when \emph{every} reachable $\tup{s^{i}_{0},\mu_i}$ is a LTI system. Two special cases of LTI systems~\cite{scheinerman:96:book} are
considered, namely, \emph{homogenous} (H) systems, where $\matr{b}$ $=$ $\matr{0}$,  and \emph{non--homogenous} (NH) systems,
where $\matr{A}$ $=$ $\matr{0}$. \textsc{Convoys} and \textsc{Non Linear Car} instances were
taken from~\cite{scala:16:ecai}. For two of the domains, namely \textsc{Agile Robot} and \textsc{Orbital Rendezvous}, no
\PDDLplus~formulation was available for us to use, so we coded them up from scratch following L\"{o}hr's description in his PhD dissertation~\citeyear{lohr:14:dissertation}.

We next briefly discuss some interesting features of the domains and how events and durative actions were handled when
present in existing \PDDLplus~formulations. In McDermott's \textsc{Convoys} the number of possible modes is given by the possible combinations of
assignments of convoys and point-to-point routes in the navigation graph. This potential combinatorial explosion was
something intended, as McDermott's points in his paper~\citeyear{mcdermott:03:icaps}. \textsc{Orbital Rendezvous} required us to compile away the durative
actions present in L\"{o}hr's model following the discussion in Section 4 of Fox \& Long paper on \PDDLplus~\citeyear{fox:pddl_plus}, but instead of introducing a ``stop``
action, we introduce an auxiliary ``timer'' variable set by the ``start'' action, whose value decreases as an effect of the process being triggered.
Besides L\"{o}hr's five hand crafted instances, we have generated a number of random ones, following a Gaussian distribution around
the initial states featured in L\"{o}hr's instances. We report the results of the planners on each set to show the robustness of the planners
when slight perturbations are introduced in initial states.
Last, the instances of \textsc{Agile Robot} we consider have no obstacles, but do have dead--ends: when (if) the robot velocity
reaches the value of $0$, there is no way to put it back into motion. Events in the domains \textsc{Powered Descent} and
\textsc{Dino's Car} were used to introduce \emph{dead end} states when the trajectories induced by plans hit a region of
the state space where a given property $\phi$ holds. We translated these directly as global constraints
 $\neg \phi$, doing away with no longer necessary auxiliary predicates present in Piotrowski's model that enable actions and processes. Conversely, global
constraints $\psi$ in \textsc{Zermelo}, \textsc{Agile Robot}, \textsc{Orbital Rendezvous} and \textsc{Non Linear Car} were translated
into events with precondition $\neg \psi$ and effects that delete irrevocably auxiliary predicates in the preconditions of actions and processes.
FSTRIPS and \PDDLplus~domains and sample instances\footnote{Available from the authors after request.}.

\begin{table*}[ht!]
	{\footnotesize
		\begin{center}
			\begin{tabular}{@{ }c@{ }c@{ }c@{ }|@{ }c@{ }c@{ }c@{ }|@{ }c@{ }c@{ }c@{ }|@{ }c@{ }c@{ }c@{ }|@{ }c@{ }c@{ }c@{ }|@{}c@{ }c@{ }c@{ }|@{}c@{ }c@{ }c@{ }}
				\input{experiments/Coverage_Tech_Report.tex}

			\end{tabular}
		\end{center}
		\caption{{\footnotesize Coverage of \UPMurphi, \Dino, \ENHSP and \FSPlus\ over the domains in Table~\ref{table:domain_taxonomy}. Timeout (TO) was set to
				$1,800$ seconds and physical memory limited to $4$ GBytes. $\Delta_{max}$ is the planning time step used,
				$I$ is the number of instances for each domain, $S$ is the number of instances
				solved by the planner, $R$ is the average run-time (in seconds) and $D$ is the average duration of the plans in seconds
				for all the domains but \textsc{Convoys}, where it is given in hours. \UPMurphi, \Dino and \ENHSP~all use $\Delta_{max}$ as
				their discretisation step. Two variants of \FSPlus\ are
%
%
%
%
				considered, \FSPlus$_{RK22}$ uses the $RK22$ integrator checking for zero crossings of Equation~\ref{eq:steady interval_invariant}, \FSPlus$_{RK22,NV}$
				uses the same integrator but does not check for zero crossings. $\Delta z$ is set to $0.1 \Delta_{max}$ for the \FSPlus\ planners. NS indicates that
				the planner did not support the domain, ``Crashes'' that the planner crashed during execution.}}
		\label{table:coverage}
	}
\end{table*}

We have compared \FSPlus\ with four state--of--the--art hybrid planners, \UPMurphi\ \cite{dellapenna:09:icaps}, \Dino\ \cite{piotrowski:16:ijcai},
\ENHSP\ \cite{scala:16:ecai}\footnote{In every case, we use the latest version available from the author's website.}
and \SMTPlan\ \cite{cashmore:16:icaps}. Table~\ref{table:coverage} shows coverage and run times for the three first planners. We have only managed to run
\SMTPlan\ on \textsc{Intercept}, where it solved all the instances in less than
a second. Its authors~reported, via personal communication, that dynamics other than non--homogenous linear ($NH$) are not supported at the time of writing this. In \textsc{Convoys} \SMTPlan\ failed to generate
a formula. 
We used the planning step $\Delta_{max}$ as the discretization parameter for \UPMurphi\, \Dino and \ENHSP. 
The values for $\Delta_{max}$ follow either from the values used in the papers discussing each domain, in the
case of \textsc{Zermelo} we performed an analysis similar to the one perfiormed by Piotrowski's 
for \textsc{Powered Descent}~\citeyear{piotrowski:16:ijcai}. In addition, \Dino\ requires bounds on plan duration
to be supplied as a parameter for its heuristic estimator. We used the values given by Piotrowski et al when available, otherwise we used the 
duration obtained by the \FSPlus\ planner. We show in Table~\ref{table:coverage} two of out of six configurations of \FSPlus\ tested, where we considered
each of the three integration methods implemented, turning on and off the testing for zero crossings. The two configurations chosen showed the best
trade--off between run times, coverage and plan duration \emph{over these benchmarks}. In all cases we used a generic implementation of Greedy Best First Search.

Table~\ref{table:coverage} shows that the two best performing \FSPlus\ configurations 
clearly dominate \UPMurphi\ and \Dino\ on every domain
but \textsc{Powered Descent}. We consider this results to be \emph{indicative}, we have observed the
performance of \UPMurphi\ and \Dino\ to be very sensitive to the parameters used to discretise time
and any \emph{numeric variables}. Compared with \ENHSP, we can see that by doing considerably more computation while searching for plans, the \FSPlus\ planners
have similar or superior runtimes and coverages 
across all the domains considered. Also, \FSPlus\ plans 
reach goal states faster
when on--line zero crossing detection is enabled. This provides evidence that the numerical integration and on--line discretization
mechanisms proposed in this planner can pay off over a broad class of domains. We also note that the \FSPlus\ planners can solve the same domains by increasing
$\Delta_{max}$, doing less work to find better plans in \textsc{Agile Robot}, but in three
instances where our dynamic discretization strategy results in very deep searches, and solving all instances of \textsc{Intercept}. Somewhat surprisingly, $\FSPlus_{RK22}$ solves less problems than  $\FSPlus_{RK22,NV}$
in \textsc{Agile Robot}, when $\Delta_{max}$ $=$ $1$.  Testing for zero crossings not only avoids 
generating unsound plans, sometimes resulting on the search abandoning a very deep plan prefix, in which case  $\FSPlus_{RK22}$ finds fewer plans as it times out. On other occasions, it does create opportunities to 
reach a goal state by interrupting an interval $H_i$, bringing about shallower searches, in which case $\FSPlus_{RK22}$ finds more plans. The latter follows from considering zero crossings for the goal formula in Equation~\ref{eq:steady interval_invariant} and preventing the planner from overshooting and ruling out
plan prefixes which result in ``orbiting'' the goal region. The most dramatic impact of this observation
becomes apparent in the \textsc{Intercept} when $\Delta_{max}$ is $1$. We consider this to be a
strength of our approach, and also highlights the limitations of the classic search method used,
Greedy Best First Search.

Validation of hybrid plans is still problematic, and warrants further attention from the community. The well--known hybrid plan validator \VAL~\cite{howey:05:ictai}  does not handle most
of the types differential equations listed in Table~\ref{table:domain_taxonomy}. Of the domains listed there, 
we could only use \VAL\ on Intercept. In the rest, we verified offline the plans generated 
by \FSPlus\ with our own validator, which is \FSPlus\ itself loading up a given plan like \VAL\ does 
and executing it using the most precise integrator available, the Implicit Euler method in
Equation~\ref{eq:implicit_euler} and the smallest $\Delta_{max}$ allowed by the precision of 
the floating point representation used, $0.001$. This value follows from multiplying by $100$ the smallest
value of $\Delta z$ for which the \texttt{C++} type \texttt{float} produces reliable results,$~1 e^{-5}$. We found no invalid plans for the $\FSPlus_{RK22}$
configuration. For $\FSPlus_{RK22,NV}$ configuration, that does not perform any checks for zero crossings
for the invariant in Equation~\ref{eq:steady interval_invariant}, we did indeed detect invalid plans
being generated in every of the domains considered.
\begin{figure*}[ht!]
	\centering
	\includegraphics[width=1.0\linewidth,keepaspectratio]{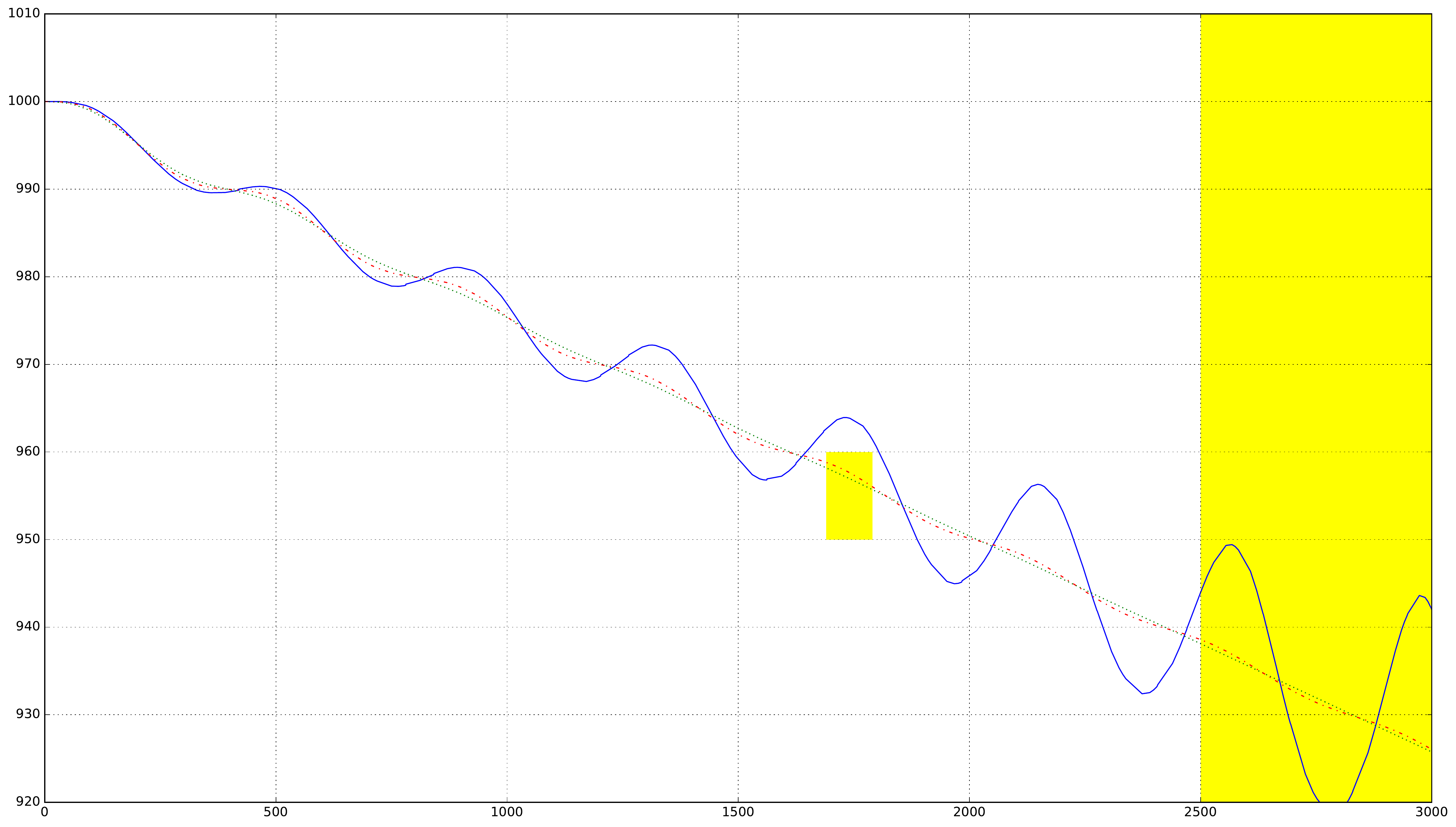}
	\caption{ Convergence of integration methods on the \textsc{Sailplane} domain, that implements the \emph{phugoids} model of glider flight. Lines represent the trajectory that follows from the plans considered during search. Trajectories induced by Explicit Euler (solid line), $RK22$ (dashed) and Implicit Euler (dotted) is shown.  See \url{https://goo.gl/SdwafF} for details and the text for discussion. $x$-axis indicates distance from the origin and the $y$-axis is the height of the glider, both measured in meters. Curves represent the trajectory of the glider as \emph{predicted} by each numerical
	integration method.}
	\label{fig:glider_integrators}
\end{figure*}
The challenge posed by arbitrary, non--trivial non--linear dynamics is illustrated next. 
Figure~\ref{fig:glider_integrators} depicts the trajectories obtained with the numerical integrators discussed 
in Section V over a domain that makes apparent
the trade-off between good convergence properties of the numerical integrator and run 
time. $\Delta z$ is set to $0.5$ in all three cases. We note that successor generation
of \UPMurphi, \Dino\ and \ENHSP\ can be interpreted as using the Explicit Euler method directly with fixed $\Delta z$ $=$ $\Delta_{max}$. When the set of goal states is the larger rectangle on the right hand side of
Figure~\ref{fig:glider_integrators} the three integrators discussed seem equivalent: they all result in trajectories that reach the goal. But, when the
set of goal states is the smaller rectangle, none of these planners would ``see'' the goal without considerably reducing the rates at which time is discretized. While the explicit Euler method requires trivial amounts of
computation, the trajectories obtained by the Runge--Kutta and Implicit Euler methods in Figure~\ref{fig:glider_integrators} are significantly more expensive, with the Implicit Euler method being
about $100$ times slower than the Runge--Kuta method we have implemented, which in turn is $3$ times slower
than the Explicit Euler method.

%% file: experiments/Coverage_Tech_Report.tex
&&& \multicolumn{3}{|c|}{\UPMurphi} & \multicolumn{3}{|c|}{\Dino} & \multicolumn{3}{|c|}{\ENHSP} & \multicolumn{3}{|c|}{\FSPlus$_{RK22,NV}$} & \multicolumn{3}{|c|}{\FSPlus$_{RK22}$} \\
 \hline  Domain & $\Delta_{max}$ &I&S&R&D&S&R&D&S&R&D&S&R&D&S&R&D\\
 \hline\multicolumn{1}{l|}{\textsc{Agile  Robot}} &0.5&100&82&1.11&14.46&82&21.10&18.68&100&10.55&8.5&100&0.47&7.0&100&8.00&6.5\\
 \multicolumn{1}{l|}{\textsc{}} &1.0&100&0&n/a&n/a&0&n/a&n/a&0&n/a&n/a&100&0.14&10.0&97&0.89&7.4\\
\multicolumn{1}{l|}{\textsc{Convoys}} &1&6&2&2.58&18.50&3&99.31&15.33&6&4.21&11.7&6&3.57&12.3&6&4.78&11.6\\
 \multicolumn{1}{l|}{\textsc{Dino  Car}} &1&10&10&1.78&15.00&10&176.00&16.80&10&0.07&12.0&10&0.01&11.0&10&0.03&11.0\\
  \multicolumn{1}{l|}{\textsc{Intercept}} &0.1&10&12&204.16&62.17&0&n/a&n/a&10&5.68&10.6&8&39.00&10.3&10&28.28&10.4\\
 \multicolumn{1}{l|}{\textsc{}} &1&10&0&n/a&n/a&0&n/a&n/a&0&n/a&n/a&0&n/a&n/a&10&85.93&10.5\\
  \multicolumn{1}{l|}{\textsc{Non  Linear  Car}} &1&8&10&1.78&15.00&10&176.00&16.80&8&0.55&10.4&8&1.68&10.5&8&0.72&10.2\\
  
 \multicolumn{1}{l|}{\textsc{Orb.  Rend. (L\"{o}hr)}} &100&5&1&35.92&9900&0&n/a&n/a&5&0.36&2300.0&5&0.26&2900.0&5&0.46&2223.4\\
 \multicolumn{1}{l|}{\textsc{ (Random)}} &100&100&1&34.20&2100&0&n/a&n/a&100&0.31&2104.0&100&0.09&2534.0&100&0.34&2059.8\\
 \multicolumn{1}{l|}{\textsc{Powered  Descent}} &1&20&10&209.64&26.10&18&47.25&40.33&20&1.99&31.4&19&2.75&31.1&16&1.80&28.5\\
  \multicolumn{1}{l|}{\textsc{Zermelo}} &100&100&0&n/a&n/a&0&n/a&n/a&92&6.16&2284.8&98&1.04&2225.5&94&1.23&2098.9\\

%% file: discussion.tex
To sum up, this paper develops formally and practically
the notions of on--line numerical integration and dynamic discretization in
the context of hybrid planning. The resulting planner, \FSPlus, seeks plans
by means of heuristic forward search, and is shown to
perform robustly and compare favorably with existing similar
planners, over a varied set of benchmarks. 
We propose alternative semantics for hybrid planning, but the purpose of doing so is 
not to supersede \PDDLplus, but rather, to offer a complementary and useful 
view on the problem, compatible with existing tools to a great degree.
We hope this work will help to draw interest from the greater
classical planning community onto this very challenging and interesting
form of planning, since we think the model presented in Section V suggests
that existing techniques developed for classical propositional planning could
be adapted to operate on it. 

In this work we have only accounted partially for natural examples
motivating the use of \PDDLplus~events to model real--world control tasks, we 
look forward to incorporate events as future work, introducing syntactic restrictions
to avoid complicating even more the validation of plans. Further exploring the impact
of syntactic restrictions looks to us as a promising future line of work, which
allows to take a more positive look at the questions posed by the difficulties of validating
the plans produced than initially suggested by the somewhat discouraging Richardson's undecidability results.
First, it is obvious to us that fragments of hybrid planning that are decidable do exist.
For instance, those domains where the expressions used in the effects of processes are such that the
trajectory induced by plans for specific tasks can be mapped into the traces of Timed Automata~\cite{alur:94:theory}. That opens up the possibility that planning algorithms
can be used to solve realizability queries over such automata scaling up as the number
of state variables grows, better than existing approaches. Second, and after acknowledging that
there is a temporal planning problem in every hybrid one, it becomes apparent to us
that a significant part of global constraints and preconditions typically featured by
many domains are about the \emph{timing} of actions or, 
\emph{indirectly}, of events which result in sets of simple linear 
numeric constraints the planner needs to check for zero crossings. In that case, zero crossing checking can be done symbolically 
as first shown by Shin \& Davies' in their planner~~\TmLpSat~\citeyear{shin:05:tmlpsat}. This suggests in turn that the on-line validation 
techniques described in Section V could be specialised to deal with specific types of 
constraints in the invariant ${\cal I}(H_i)$. Both Satisfiability Modulo Theory (SMT)~\cite{barrett:08:smt}
and Constraint Programming (CP)~\cite{marriott:98:cp,stuckey:16:cp} look to us as complementary and 
ready to use frameworks facilitating such hybrid reasoning, with plenty of available solvers
to experiment with. Doing so would allow to offer more solid guarantees on the validity
of plans, at least when we have to deal with such types of constraints exclusively.

Our planner can be readily extended and is easy to interface with existing solvers and
simulators either via semantic attachments that enhance the fidelity of the domain model,
or by embedding an instance of the planner into
existing simulation software via run--time dynamic linking or statically if access to the
source code is available. Such a capability, while purely
practical, suggest that \FSPlus\ can be used to
develop intelligent and highly interactive computer--assisted design (CAD) 
tools for complex engineering
problems, significantly improving the ability to explore the design space over existing ad-hoc solutions developed for specific problems like \textsc{Nasa Jpl}'s Astrodynamics Toolkit~\cite{JAT:2013}
or general-purpose, off-the-shelf sofware packages such as \textsc{Matlab}'s \textsc{Simulink}~\cite{MATLAB:2016}. We are actively working on developing a broader set of benchmarks: in this paper we have all but considered a \emph{tiny} sample from the set of tasks that can be modelled. This will allow us to further stress \FSPlus, and to identify
stakeholders and use cases for such CAD tools.

Last, we acknowledge that some of the problems modeled by \FPDDL~ domains considered 
have been approached using the models, tools and algorithms widely used by the community of 
robotic motion and path planning. We are actively engaging with existing motion planning frameworks such as \textsc{OMPL}~\cite{sucan:12:ompl}, simulators and benchmarks, and we look forward to
bridge to some extent, the methodological and theoretical gap between ``task'' and ``motion'' planning as isolated \emph{disciplines}.
